\documentclass[10pt,twocolumn]{article}

\usepackage[T1]{fontenc}
\usepackage[utf8]{inputenc}
\usepackage{authblk}
\usepackage{arxiv}

\usepackage{graphicx}
\usepackage{amsfonts}
\usepackage{amsmath}
\usepackage{graphicx}
\usepackage{epstopdf}
\usepackage{float}
\usepackage{color}
\usepackage{relsize}
\usepackage{tikz}
\usepackage{color}
\usepackage[font=itshape, begintext=‘, endtext=’]{quoting}
\usepackage[colorlinks,bookmarksnumbered,bookmarksopen,citecolor=blue,urlcolor=blue,linkcolor=blue,]{hyperref}
\usetikzlibrary{intersections,shapes.arrows}
\usetikzlibrary{patterns, positioning, arrows,spy,shapes.geometric,calc}
\usetikzlibrary{shapes,arrows}
\usepackage{multicol}

\newcommand{\dpar}[2]{\frac{\partial #1}{\partial #2}}

\newcommand{\bs}[1]{\boldsymbol{#1}}
\newcommand{\nexp}[1]{\cdot 10^{#1}}

\usepackage{pgfplotstable}
\usepgfplotslibrary{statistics}
\pgfplotsset{
/pgfplots/custom legend/.style={
legend image code/.code={
\draw [only marks,mark=square]
plot coordinates {(0.3cm,0cm)};
}, },}

\makeatletter
\pgfplotsset{
    boxplot prepared from table/.code={
        \def\tikz@plot@handler{\pgfplotsplothandlerboxplotprepared}%
        \pgfplotsset{
            /pgfplots/boxplot prepared from table/.cd,
            #1,
        }
    },
    /pgfplots/boxplot prepared from table/.cd,
        table/.code={\pgfplotstablecopy{#1}\to\boxplot@datatable},
        row/.initial=0,
        make style readable from table/.style={
            #1/.code={
                \pgfplotstablegetelem{\pgfkeysvalueof{/pgfplots/boxplot prepared from table/row}}{##1}\of\boxplot@datatable
                \pgfplotsset{boxplot/#1/.expand once={\pgfplotsretval}}
            }
        },
        make style readable from table=lower whisker,
        make style readable from table=upper whisker,
        make style readable from table=lower quartile,
        make style readable from table=upper quartile,
        make style readable from table=median,
        make style readable from table=lower notch,
        make style readable from table=upper notch
}
\makeatother

\title{Port-metriplectic neural networks: thermodynamics-informed machine learning of complex physical systems}

\author[1]{Quercus Hern\'andez}
\author[2]{Alberto Bad\'ias}
\author[3,4]{Francisco Chinesta}
\author[1]{El\'ias Cueto}
\affil[1]{{\small Aragon Institute of Engineering Research (I3A). University of Zaragoza. Zaragoza, Spain.}}
\affil[2]{{\small Higher Technical School of Industrial Engineering, Polytechnic University of Madrid. Madrid, Spain.}}
\affil[3]{{\small ESI Group chair. PIMM Lab. ENSAM Institute of Technology. Paris, France.}}
\affil[4]{{\small CNRS@CREATE LTD. Singapore.}}

\begin{document} 

\twocolumn[
\maketitle

\begin{abstract}
We develop inductive biases for the machine learning of complex physical systems based on the port-Hamiltonian formalism. To satisfy by construction the principles of thermodynamics in the learned physics (conservation of energy, non-negative entropy production), we modify accordingly the port-Hamiltonian formalism so as to achieve a port-metriplectic one. We show that the constructed networks are able to learn the physics of complex systems by parts, thus alleviating the burden associated to the experimental characterization and posterior learning process of this kind of systems. Predictions can be done, however, at the scale of the complete system. Examples are shown on the performance of the proposed technique.
\end{abstract}
\hfill \break]

%\begin{multicols}{2}

\section{Introduction}\label{sec1}

Recently, the possibility of developing learned simulators has attracted an important research activity in the computational mechanics community and beyond. By ``learned simulators'' we mean methodologies able to learn from data the dynamics of a physical system so as to perform accurate predictions about previously unseen situations without the burden associated to the construction of numerical models by means of finite elements, finite volumes or similar techniques \cite{stachenfeld2021learned,allen2022physical,battaglia2018relational}. Among their advantages we can cite that they are based on reusable architectures, can be optimized to work under really stringent real-time feedback rates, and are specially well suited for optimization and inverse problems.

While original, black-box approaches showed great promise, both industry and academia are reluctant to generalize their use, since small modifications in the input data may cause nonsense results. This is at the origin of the development and employ of inductive biases during the learning process \cite{battaglia2018relational,bhattoo2021lagrangian}. An inductive bias allows the learning algorithm to prioritize one particular solution over any other \cite{mitchell1980need}. This is particularly interesting for physical phenomena for which previous knowledge exists. Paul Dirac once said that \cite{dirac1929quantum} 
\begin{quoting}The underlying physical laws necessary for the mathematical theory of a large part of physics and the whole of chemistry are thus completely known, and the difficulty is only that the exact application of these laws leads to equations much too complicated to be soluble. 
\end{quoting} Therefore, in the presence of centuries of knowledge about virtually any physical phenomena, it is simply nonsense to ignore it and to favor theory-blind, black-box approaches.

In this paper we develop a novel strategy based on the port-Hamiltonian formalism, which we extend so as to comply with the first and second principles of thermodynamics by construction \cite{van2014port,beattie2019robust,rashad2020twenty}. Port-Hamiltonian formalisms extend the well-known Hamiltonian (thus, conservative) physics to open systems and introduce the possibility of dissipation and control through external actuation within this theory. We show here, however, that general port-Hamiltonian systems do not comply a priori with the laws of thermodynamics and modify them so as to ensure this fulfillment. Based on this new formalism, which we call port-metriplectic, since it is at the same time metric and symplectic, we construct a deep neural network methodology to learn the physics of complex systems from data. The resulting port-metriplectic networks will comply by construction with the principles of thermodynamics---that can be enforced through hard or soft constraints---while they allow to analyze complex systems by parts. These parts will then communicate through energy ports to construct the final, complex systems.

The outline of the paper is as follows. In Section \ref{TINNs} we review the state of the art in the development of machine learning strategies that impose energy conservation by using a Hamiltonian formalism. We include here neural networks based upon port-Hamiltonian formalisms, which we show not be necessarily compliant with the principles of thermodynamics. We then develop the concept of port-metriplectic networks in Section \ref{PMNNs}. Then, in Section \ref{results} we analyze the performance of the just developed neural networks, while in Section \ref{conc} we draw some conclusions.

\section{Hamiltonian neural networks}\label{TINNs}

\subsection{Reversible dynamics as an inductive bias}

Learning the physics of a given phenomenon from data can be seen as learning a dynamical problem \cite{E2017}. If we assume that the problem is governed by a set of variables $\bs z$, which we can measure experimentally---a detailed discussion on the limitations and implications of this assumption can be found at \cite{cueto2022thermodynamics}---, then the problem of learning the evolution of the system in time can be seen as finding the structure of the dynamical problem
\begin{equation}\label{eq1}
    \dot{\bs z} = \frac{d \bs z}{d t} = f(\bs z,t),\;\; \bs z(0) = \bs z_0,
\end{equation}
or, in other words, to find by regression the flow map
\begin{equation}
    \bs z_0 \rightarrow \bs z(T,\bs z_0).
\end{equation}
Equivalently, we must find the particular form of the function $f$ governing the dynamics of the system. This is done by regression, where neural networks play an important role, but by no means constitute the only possibility, as in \cite{gonzalez2019thermodynamically,gonzalez2019learning,gonzalez2021learning}.

This particular form of seeing the problem has important advantages. For instance, if the system under scrutiny is known to be conservative, or reversible, we can impose as an inductive bias the Hamiltonian form of the sought function $f$,
\begin{equation}\label{Newton}
 \dot{\bs z} = \frac{d \bs z}{d t} =  \bs L\dpar{\mathcal H}{\bs z} = \bs L\dpar{E}{\bs z},
\end{equation}
where the Hamiltonian, $\mathcal H$, whose canonical form depends on the position and momenta of particles, is now the total energy of the system, $E$. Under this prism, the problem (\ref{eq1}) is now seen as to find the precise form of the skew-symmetric (symplectic) matrix $\bs L$ and the form of the energy of the system, $E(\bs z)$. If we enforce the particular form given by Eq. (\ref{Newton}) during our regression procedure, it is straightforward to prove that the resulting evolution will be conservative.

Many works have leveraged this approach. Several authors take advantage of the Hamiltonian structure to construct symplectic integrators to predict conservative dynamical systems \cite{jin2020sympnets,chen2022learning,miller2020mastering}. Others, use the Hamiltonian principles to design more expressive deep neural network architectures \cite{galimberti2021unified} or to find the Hamiltonian function and phase space from data \cite{bertalan2019learning,toth2019hamiltonian}. The Hamiltonian paradigm is also widely used in quantum mechanics, where similar deep learning literature can be found in problems such as electron dynamics \cite{bhat2020machine}, learning ground states \cite{kochkov2021learning} or optimal control \cite{gao2022harnessing}. Alternative formulations can be developed by resorting to the equivalent Lagrangian formalism, see \cite{bhattoo2021lagrangian,lutter2019deep,zhong2020unsupervised,lee2002enhanced,allen2020lagnetvip}, among others.

\subsection{Port-Hamiltonian neural networks}

If the physical phenomenon at hand is known to be dissipative, or if the system is open and thus no guarantee on the conservation of energy exists, things become more intricate. For dissipative systems, the easiest form of the evolution Eq. (\ref{eq1}) could be, perhaps, a gradient flow \cite{hohenberg1977theory}. Their evolution can be established after some (dissipation) potential $\mathcal R$ in the form \cite{CiCP-28-1639}
$$
\dot{\bs z}= -\frac{\partial \mathcal R}{\partial \bs z}.
$$

Recently, the so-called Symplectic ODE nets (symODEN) \cite{zhong2020dissipative,zhong2021benchmarking}, have tackled the issue of introducing dissipation in the learned description of the physics. It is also the approach followed in \cite{gruver2022deconstructing}. More recently, two distinct works have tackled the dissipation problem by relaxing equivariance in the networks \cite{han2022learning,wang2022approximately}.

These works seem to be closely related to the vast corps of literature on the port-Hamiltonian approach to dynamical systems  \cite{van2014port,rashad2020twenty,beattie2019robust}. Port-Hamiltonian systems assume an evolution of the system in the form
\begin{equation}\label{pH}
\begin{bmatrix}
\dot{\bs q} \\ \dot{\bs p}
\end{bmatrix} =
\left(
\begin{bmatrix}
\bs 0 & \bs I \\
-\bs I & \bs 0
\end{bmatrix}-\bs D(\bs q) 
\right)
\begin{bmatrix}
\frac{\partial \mathcal H}{\partial \bs q} \\ \frac{\partial \mathcal H}{\partial \bs p}
\end{bmatrix} + 
\begin{bmatrix}
\bs 0\\ \bs g(\bs q)
\end{bmatrix}\bs u,
\end{equation}
where $(\bs{q},\bs{p})$ are the generalized position and momenta, dissipation is included by adding a symmetric, positive semi-definite matrix $\bs D$, and control is considered through an actuation term $\bs u$ and a non-linear function of the position $\bs{g}(\bs{q})$. Eq. (\ref{pH}) reduces to the Hamiltonian description if no dissipation nor control are considered. Here, we have assumed a canonical form for the Hamiltonian, i.e., that it depends on a set of variables $\bs z= \{\bs q,\bs p\}$. More general forms can be expressed similarly.

The true advantage of using port-Hamiltonian formalisms as inductive biases in the learning procedure stems from the fact that, on one side, they allow the introduction of dissipation and control and, on the other, they model open systems (as opposed to classical Hamiltonian descriptions where energy conservation assumes inherently that the system is closed) \cite{eidnes2022port}.

Therefore, the use of port-Hamiltonian formalisms as inductive biases in learning processes is extremely interesting. However, as will be demonstrated in the next section, classical port-Hamiltonian schemes do not guarantee a priori to comply with the laws of thermodynamics, see \cite{cueto2022thermodynamics}.

\section{Port-metriplectic neural networks}\label{PMNNs}

\subsection{Metriplectic biases for dissipative phenomena}

In the case of dissipative phenomena, the first in proposing the introduction of a second potential, the so-called Mathieu potential, seems to have been Morrison \cite{morrison1984bracket,morrison1986paradigm}, Grmela \cite{grmela1984particle,grmela1984bracket} and Kaufman \cite{kaufman1984dissipative}. They suggested to consider an evolution of the governing variables of the type
 \begin{equation}\label{GENERIC2}
\dot{\bs z} = \bs L(\bs z) \frac{\partial E}{\partial \bs z} + \bs M(\bs z)\frac{\partial S}{\partial \bs z},
\end{equation}
where $S$ is precisely this second (dissipation) potential, entropy.

This formulation is often referred to as metriplectic, since it is metric and symplectic at the same time. Here, $\bs M(\bs z)$ is a symmetric, positive semi-definite dissipation matrix and $\bs L(\bs z)$, the Poisson matrix, continues to be skew-symmetric.

However, for this formulation to be consistent with the principles of thermodynamics, two additional conditions must hold, the so-called degeneracy conditions:
\begin{equation}\label{deg1}
\bs L(\bs z)\frac{\partial S}{\partial \bs z} = \bs 0,
\end{equation}
and
\begin{equation}\label{deg2}
\bs M(\bs z)\frac{\partial E}{\partial \bs z} = \bs 0,
\end{equation}
which give rise to the General Equation for the non-Equilibrium Reversible-Irreversible Coupling, GENERIC, equations \cite{ottinger1997dynamics,ottinger2005beyond,grmela2018generic,grmela2019gradient,pavelka2018multiscale}. 

In a nutshell, Eqs. (\ref{deg1}) and (\ref{deg2}) state that the energy potential is independent of dissipation, whereas entropy is unrelated to the energy conservation. If they hold, it is straightforward to demonstrate that, given the skew-symmetry of $\bs L$,
$$
\dot{E}(\bs z) = \frac{\partial E}{\partial \bs z}\dot{\bs z} =0,
$$
and
$$
\dot{S} = \frac{\partial S}{\partial \bs z}\dot{\bs z} = \frac{\partial S}{\partial \bs z}\bs M(\bs z) \frac{\partial S}{\partial \bs z}\geq 0,
$$
given the positive semi-definiteness of $\bs M$.

These properties have been leveraged in some of our former works to develop what we have coined as thermodynamics-informed neural networks \cite{hernandez2021structure,hernandez2021deep,hernandez2022thermodynamics}.

Given experimental data sets $\mathcal{D}_i$ containing labelled pairs of a single-step state vector $\bs{z}_t$ and its evolution in time $\bs{z}_{t+1}$,
\begin{equation}
\mathcal{D}=\{\mathcal{D}_i\}_{i=1}^{N_{\text{sim}}},\quad\mathcal{D}_i =\{(\bs{z}_t,\bs{z}_{t+1})\}_{t=0}^{T},
\end{equation}
we construct a neural network by considering two different loss terms. First, a data-loss term that takes into account the correctness of the network prediction of the state vector at subsequent time steps by integrating GENERIC in time, i.e.,
\begin{equation*}
\mathcal{L}^{\text{data}}_n=\left\Vert\dot{\bs{z}}^{\text{GT}}_n-\dot{\bs{z}}^{\text{net}}_n\right\Vert^2_2,
\end{equation*}
with $\Vert\cdot\Vert_2$ the L2-norm, $\dot{\bs{z}}^{\text{GT}}_n$ is ground truth solution and $\dot{\bs{z}}^{\text{net}}_n$ is the network prediction. The choice of the time derivative instead of the state vector itself is employed to regularize the global loss function to a uniform order of magnitude with respect to the degeneracy terms.

We then consider a second loss term to take into account the fulfillment of the degeneracy equations in a soft way,
\begin{equation*}
\mathcal{L}^{\text{deg}}_n=\left\Vert\bs{L}\dpar{S}{\bs{z}_n}\right\Vert^2_2+\left\Vert\bs{M}\dpar{E}{\bs{z}_n}\right\Vert^2_2.
\end{equation*}

These networks have have demonstrated to work very well for physics perception and reasoning in combination with computer vision \cite{moya2021physics,moya2022physics}. 

These two loss terms are weighted and averaged over the $N_{\text{batch}}$ batched snapshots.
\begin{equation}
\mathcal{L}=\frac{1}{N_{\text{batch}}}\sum_{n=0}^{N_{\text{batch}}}(\lambda \mathcal{L}^{\text{data}}_n+\mathcal{L}^{\text{deg}}_n).
\end{equation}

Alternative formulations of these thermodynamics-informed networks exist in which the degeneracy conditions are imposed in hard way, see \cite{zhang2021gfinns} and \cite{lee2021machine}.

It is worth noting that, by comparing Eqs. (\ref{GENERIC2}) and (\ref{deg1})-(\ref{deg2}), on one side, and Eq. (\ref{pH}), on the other, one readily concludes that port-Hamiltonian biases do not necessarily ensure the fulfillment of the principles of thermodynamics. Note that, since entropy does enter the classical port-Hamiltonian formulation, it is difficult to impose the fulfillment of the second principle of thermodynamics. Therefore, we suggest to extend the GENERIC formalism to open systems so as to develop alternative port-metriplectic biases. These are developed in the next section.

\subsection{Port-metriplectic neural networks}

\begin{figure*}[ht]
\centerline{\includegraphics[width=\textwidth]{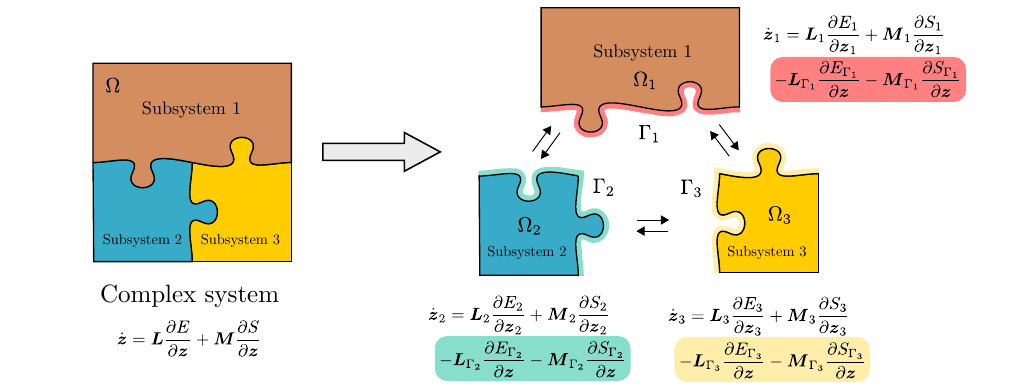}}
\caption{Complex system created as a connected group of subsystems $\Omega_i$. The dynamics of any subsystem is described using a GENERIC formulation including conservative and dissipative terms, taking also into account the external contributions in the boundary terms $\Gamma_i$. Sub-systems communicate between each other through ports by the exchange of energy and entropy.\label{fig:GENERICscheme}}
\end{figure*}

Very few works exist, to the best of our knowledge, on the development of GENERIC formulations for open systems, that may lead to the development of port-metriplectic formulations. Maybe the only exception is \cite{PhysRevE.73.036126}, later on revisited by \cite{badlyan2018open,betsch2018variational}, both published in conference proceedings and, of course, with no machine learning approximations. Both approaches are essentially identical, and start from the bracket formulation of GENERIC. Operators $\bs L$ and $\bs M$ define a bracket structure of the type
\begin{equation}\label{eq:brackets}
\dot{\bs z}=\lbrace\bs{z},E\rbrace+[\bs{z},S],
\end{equation}
where $\lbrace\cdot,\cdot\rbrace$ is the so-called Poisson bracket and $[\cdot,\cdot]$ represents the dissipative bracket \cite{ottinger1997dynamics,grmela1997dynamics,pavelka2018multiscale}.

For open systems, these brackets take the form
\begin{equation}
    \lbrace\cdot,\cdot\rbrace = \lbrace\cdot,\cdot\rbrace_{\text{bulk}} + \lbrace\cdot,\cdot\rbrace_{\text{boun}},
\end{equation}
and
\begin{equation}
    [\cdot,\cdot] = [\cdot,\cdot]_{\text{bulk}} + [\cdot,\cdot]_{\text{boun}}.
\end{equation}
In other words, both brackets are decomposed additively into bulk and boundary contributions. With this decomposition in mind, the GENERIC principle (\ref{GENERIC2}) now reads
\begin{multline}\label{PM}
    \dot{\bs z} = \lbrace \bs z,E\rbrace_{\text{bulk}} + [\bs z,S]_{\text{bulk}} \\
    = \lbrace \bs z,E\rbrace + [\bs z,S] - \lbrace \bs z,E\rbrace_{\text{boun}} - [\bs z,S]_{\text{boun}}.
\end{multline}
The degeneracy conditions (\ref{deg1}) and (\ref{deg2}) must be satisfied by the bulk operators only, since it is possible, in general, that there may be a reversible flux of entropy at the boundary or, equivalently, an irreversible flux of energy at the boundary \cite{PhysRevE.73.036126},
\begin{equation}\label{deg12}
\bs L_{\text{bulk}}(\bs z)\frac{\partial S_{\text{bulk}}}{\partial \bs z} = \bs 0,
\end{equation}
and
\begin{equation}\label{deg22}
\bs M_{\text{bulk}}(\bs z)\frac{\partial E_{\text{bulk}}}{\partial \bs z} = \bs 0,
\end{equation}
The particular form of the boundary terms in Eq. (\ref{PM}) depends, of course, of the particular phenomenon under scrutiny, but in a general way it can be expressed using $\bs L$ and $\bs M$ operators as
 \begin{align}\nonumber
\dot{\bs z} & = \bs L \frac{\partial E}{\partial \bs z} + \bs M\frac{\partial S}{\partial \bs z} \\ & - \bs L_{\text{boun}} \frac{\partial E_{\text{boun}}}{\partial \bs z} - \bs M_{\text{boun}}\frac{\partial S_{\text{boun}}}{\partial \bs z}.\label{eq:port}
\end{align}
More particular expressions can be developed if we know in advance some properties of the system at hand. For instance, in Section \ref{pendu} we deal with a double pendulum by learning the behavior of each pendulum separately. If we know in advance that the only boundary term comes from the energy-entropy pair transmitted by the other pendulum, and no other external contribution is present, more detailed assumptions in the form of degeneracy conditions can be assumed. This may lead to a decrease in learning time or the employ of less data.

Fig. \ref{fig:GENERICscheme} sketches the approach developed herein for complex systems. In the numerical results section below we explore the particular form that these terms could acquire for both finite and infinite dimensional problems.

We propose two learning procedures which correspond to different level of information available of the dynamics of the system. In the first example, we focus on two coupled subsystems in which we learn the self and boundary contributions of both subsystems to the global dynamics of the problem. This is the case when the interest is focused on the complete system divided into smaller subsystems. In the second example, we suppose that the external influence is determined by a load vector as a result of an unknown external interaction with another subsystem. Thus, the learning procedure is focused on the self and boundary contributions of only one subsystem based on an external interaction. This case is convenient for applications where only partial information of the system is available.

\section{Numerical results}\label{results}

\subsection{Double thermoelastic pendulum}\label{pendu}

\begin{figure}[h]
\centerline{
\begin{tikzpicture}

% Origen
\node[draw,circle,scale=0.25*1.5] (origin) at (0,0) {};
\node [fill,pattern=north east lines,draw=none,minimum width=0.8cm*1.5,minimum height=0.2cm*1.5,anchor=south] (ground) at ($(origin) + (0,0.15*1.5)$)  {};
\draw[very thick] (-0.4*1.5,0.15*1.5) -- (0.4*1.5,0.15*1.5);
\node (ground_left) at ($(ground) + (0.3,-0.02)$) {};
\node (ground_right) at ($(ground) + (-0.3,-0.02)$) {};
\draw (ground_left) -- (origin);
\draw (ground_right) -- (origin);

% Coords
\node (x) at (1,0) {$\bs{x}$};
\node (y) at (0,-1) {$\bs{y}$};
\draw[-stealth] (origin)  -- (x);
\draw[-stealth] (origin)  -- (y);

% Bolas 1 y 2
\node[draw,thick,circle,scale=1.5] (ball_1) at (1*1.5,-1.5*1.5) {};
\node[draw=none,below=.25cm*1.5] at (ball_1) {$m_1$};
\node[draw=none] (spring_1) at (0.1,-1*1.5) {$\lambda_1, C_1$};
\node[draw,thick,circle,scale=1.5] (ball_2) at (3.5*1.5,-3*1.5) {};
\node[draw=none,above=.25cm*1.5] at (ball_2) {$m_2$};
\node[draw=none] (spring_2) at (2.5*1.5,-1.9*1.5) {$\lambda_2, C_2$};

% Velocidades
\node (p_1) at ($(ball_1) + (0.5*1.5,1*1.5)$) {$\bs{p}_1$};	
\node (p_2) at ($(ball_2) + (-1.5*1.5,0.3*1.5)$) {$\bs{p}_2$};	
\draw[-stealth] (ball_1) -- (p_1);
\draw[-stealth] (ball_2) -- (p_2);

% Muelles
\draw[decorate,decoration={aspect=0.5, segment length=1mm, amplitude=1mm,coil}] (origin) -- (ball_1);
\draw[decorate,decoration={aspect=0.5, segment length=1.5mm, amplitude=1mm,coil}] (ball_1) -- (ball_2);

\end{tikzpicture}
}
\caption{Double thermoelastic pendulum. A single pendulum with external perturbations is learned, and the coupling between both systems is achieved via the port-metriplectic framework.\label{fig:double_pendulum}}
\end{figure}
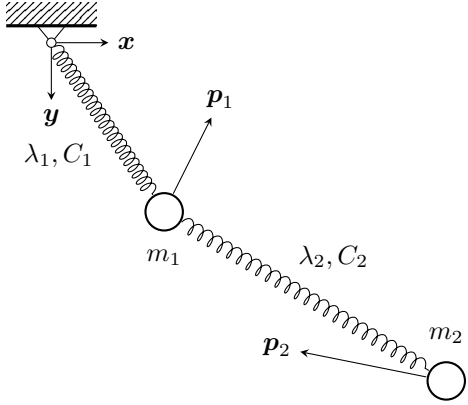

The first example is a double thermoelastic pendulum consisting of two masses $m_1$ and $m_2$ connected by two springs of variable lengths $\lambda_1$ and $\lambda_2$ and natural lengths at rest $\lambda_1^0$ and $\lambda_2^0$, as depicted in Fig. \ref{fig:double_pendulum}.

The set of variables describing each pendulum are here chosen to be
\begin{equation}\label{eq:double_z}
\mathcal{S}=\{\bs{z}=(\bs{q},\bs{p},s)\in(\mathbb{R}^2\times\mathbb{R}^2\times\mathbb{R})\}.
\end{equation}
where $\bs{q}$, $\bs{p}$ and $s$ are the position, linear momentum and entropy of the pendulum mass.

The evolution of the state variables of the second pendulum is defined as
 \begin{align*}
\dot{\bs z}_2 & = \bs{L}_2 \frac{\partial E_2}{\partial \bs z_2} + \bs{M}_2\frac{\partial S_2}{\partial \bs z_2} - \bs M_{\text{boun},2}\frac{\partial S_{\text{boun},2}}{\partial \bs z_2},
\end{align*}
where the first two positive terms describe the self contribution of the simple pendulum (conservative and dissipative effects) and the third term describes the dissipative effect produced by the first pendulum affecting over the second pendulum.

On the other hand, the evolution of the state variables of the first pendulum is defined as
 \begin{multline}\label{sep}
\dot{\bs z}_1  = \bs{L}_1 \frac{\partial E_1}{\partial \bs z_1} + \bs{M}_1\frac{\partial S_1}{\partial \bs z_1} \\  - \bs L_{\text{boun},1} \frac{\partial E_{\text{boun},1}}{\partial \bs z_1} - \bs M_{\text{boun},1}\frac{\partial S_{\text{boun},1}}{\partial \bs z_1},
\end{multline}
where in this case the first two positive terms describe the self contribution of the first simple pendulum (conservative and dissipative effects) and the third and fourth terms describe the external contribution on the conservative and dissipative parts, both produced by the influence of the second pendulum over the first pendulum.

Note that the first pendulum has no conservative contribution to the second pendulum, i.e., the term
$$
\bs L_{\text{boun},2} \frac{\partial E_{\text{boun},2}}{\partial \bs z_2}
$$
does not exist. However, there is a conservative contribution from the second pendulum on the first pendulum, see \cite{romero2009thermodynamically}.

It is worth noting, as previously pointed out in \cite{eidnes2022port}, that the fact that every term in Eq. (\ref{sep}) depends on the state variables ${\bs z}_1$ makes the learning procedure more intricate. This is caused by the non-separable structure of Eq. (\ref{sep}). This problem is not present if the port terms depend only on time, as it is the case in Section \ref{vigas} below. To overcome this limitation, we employ a structure-preserving neural network for each of the terms in Eq. (\ref{sep}). These networks share the weights, however, for both pendula, if they are known in advance to be identical.

The fact of using individual approximations of the dynamics of each subsystem (each pendulum) allows to use artificial neural networks of considerably smaller size with respect to an analysis of the whole problem using a larger number of variables to describe the global state \cite{hernandez2021structure}.

The database consists of 50 different simulations with random initial conditions of position $\bs{q}$ and linear momentum $\bs{p}$ of both masses $m_1$ and $m_2$ around a mean position and linear momentum of $\bs{q}_1=(4.5,\;4.5)$ m, $\bs{p}_1=(2,\;4.5)$ kg$\cdot$m/s, and $\bs{q}_2=(-0.5,\;1.5)$ m, $\bs{p}_2=(1.4,\;-0.2)$ kg$\cdot$m/s respectively. The masses of the double pendulum are set to $m_1 = 1$ kg and $m_2=2$ kg, joint with springs of a natural length of $\lambda^0_1=2$ m and $\lambda^0_2=1$ m and thermal constant of $C_1=0.02$ J and $C_2=0.2$ J and conductivity constant of $\kappa = 0.5$. Note that the double pendulum constitutes a closed system as a whole, but this is not the case for each one of the simple pendula. Both start from a temperature of 300K. The simulation time of the movement is $T = 60$ s in $N_T=200$ time increments of $\Delta t = 0.3$ s.

\begin{figure}[h]
\centering
\pgfplotstableread{results/double_box/error_double_train.txt}\errortrain
\pgfplotstableread{results/double_box/error_double_test.txt}\errortest

\begin{tikzpicture}
\pgfplotsset{width=\textwidth, height=6cm}

  % q
  \begin{semilogyaxis}
  [xshift=0cm,
  area legend, boxplot/draw direction=y,
  grid=major, % Display a grid
  grid style={dashed,gray!30}, % Set the style
  cycle list={{red},{blue}},
  boxplot={draw position={1/3 + floor(\plotnumofactualtype/2) + 1/3*mod(\plotnumofactualtype,2)},box extend=0.3,},
  x=1cm,xtick={0,1,2,...,10},x tick label as interval,
  xticklabels={{$\bs{q}$},{$\bs{p}$},{$s$}},
  ticklabel style={font=\scriptsize},
  ylabel={Relative L2 error},
  custom legend,legend pos=outer north east,legend cell align=left,
  legend entries = {Train, Test},
  ] 
  	% q
  	% Train
	\addplot+[boxplot prepared from table={table=\errortrain,row=0,
    lower whisker=lw,
    upper whisker=uw,
    lower quartile=lq,
    upper quartile=uq,
    median=med}, boxplot prepared]
    coordinates {};
    % Test
	\addplot+[boxplot prepared from table={table=\errortest,row=0,
    lower whisker=lw,
    upper whisker=uw,
    lower quartile=lq,
    upper quartile=uq,
    median=med}, boxplot prepared] 
    coordinates {}; 
    
    % v
    % Train
	\addplot+[boxplot prepared from table={table=\errortrain,row=1,
    lower whisker=lw,
    upper whisker=uw,
    lower quartile=lq,
    upper quartile=uq,
    median=med}, boxplot prepared] 
    coordinates {}; 
    % Test
	\addplot+[boxplot prepared from table={table=\errortest,row=1,
    lower whisker=lw,
    upper whisker=uw,
    lower quartile=lq,
    upper quartile=uq,
    median=med}, boxplot prepared] 
    coordinates {}; 

    % s
    % Train
	\addplot+[boxplot prepared from table={table=\errortrain,row=3,
    lower whisker=lw,
    upper whisker=uw,
    lower quartile=lq,
    upper quartile=uq,
    median=med}, boxplot prepared] 
    coordinates {}; 
    \addlegendentry{~Train}
    % Test
	\addplot+[boxplot prepared from table={table=\errortest,row=3,
    lower whisker=lw,
    upper whisker=uw,
    lower quartile=lq,
    upper quartile=uq,
    median=med}, boxplot prepared] 
    coordinates {}; 
    \addlegendentry{~Test}

  \end{semilogyaxis}
  
\end{tikzpicture}
\caption{Box plots for the relative L2 error for all the rollout snapshots of the double pendulum in both train and test cases. The state variables represented are position ($\bs{q}$), momentum ($\bs{p}$), and entropy ($s$).}
\label{fig:double_box}
\end{figure}
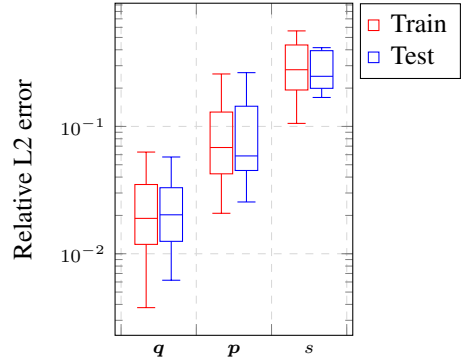

The boxplot in Fig. \ref{fig:double_box} shows the statistics of the L2 relative error of the rollout train and test simulations. 

\subsection{Interacting beams}\label{vigas}

\begin{figure}[h]
\centerline{\includegraphics[width=\linewidth]{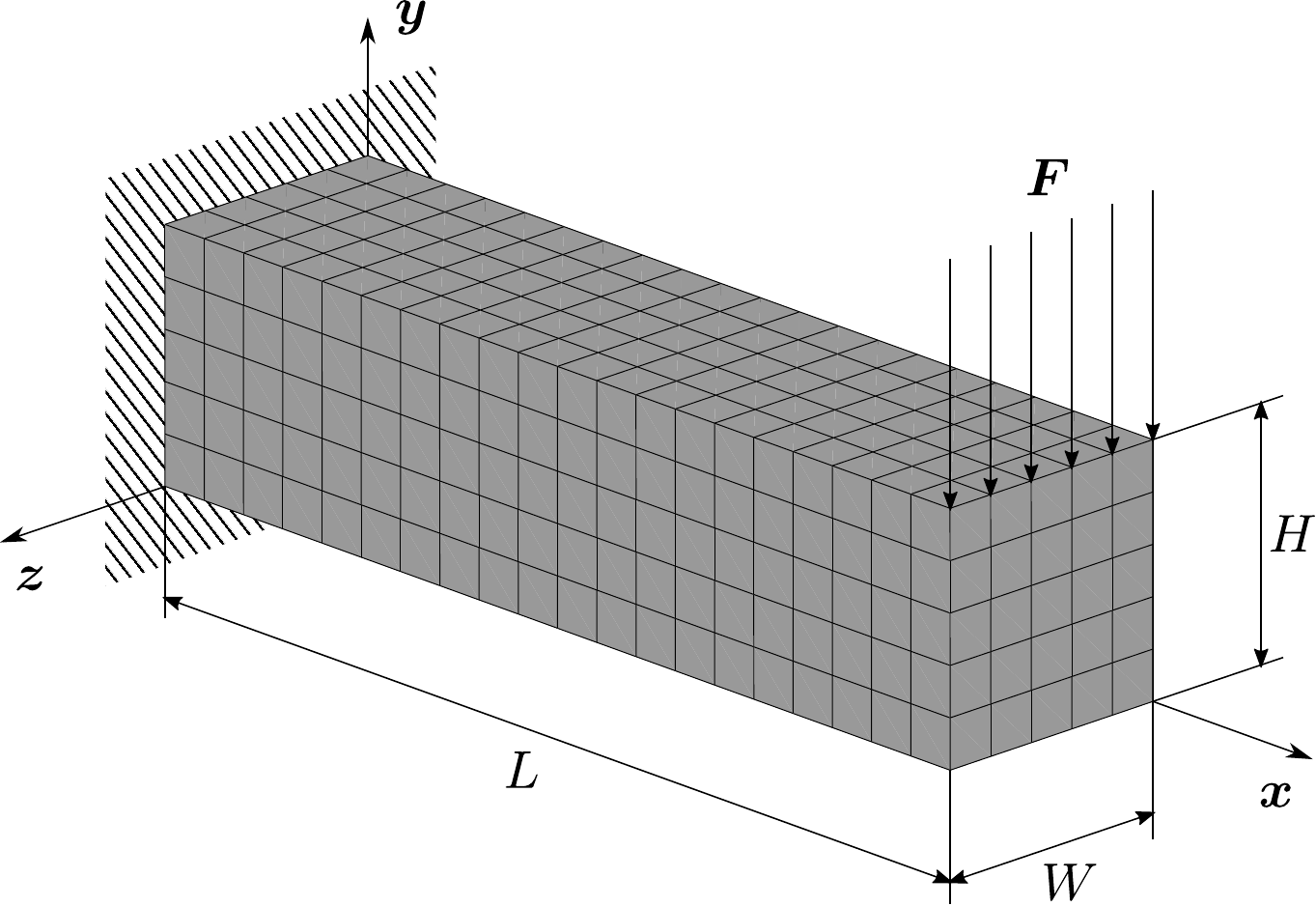}}
\centerline{\includegraphics[width=\linewidth]{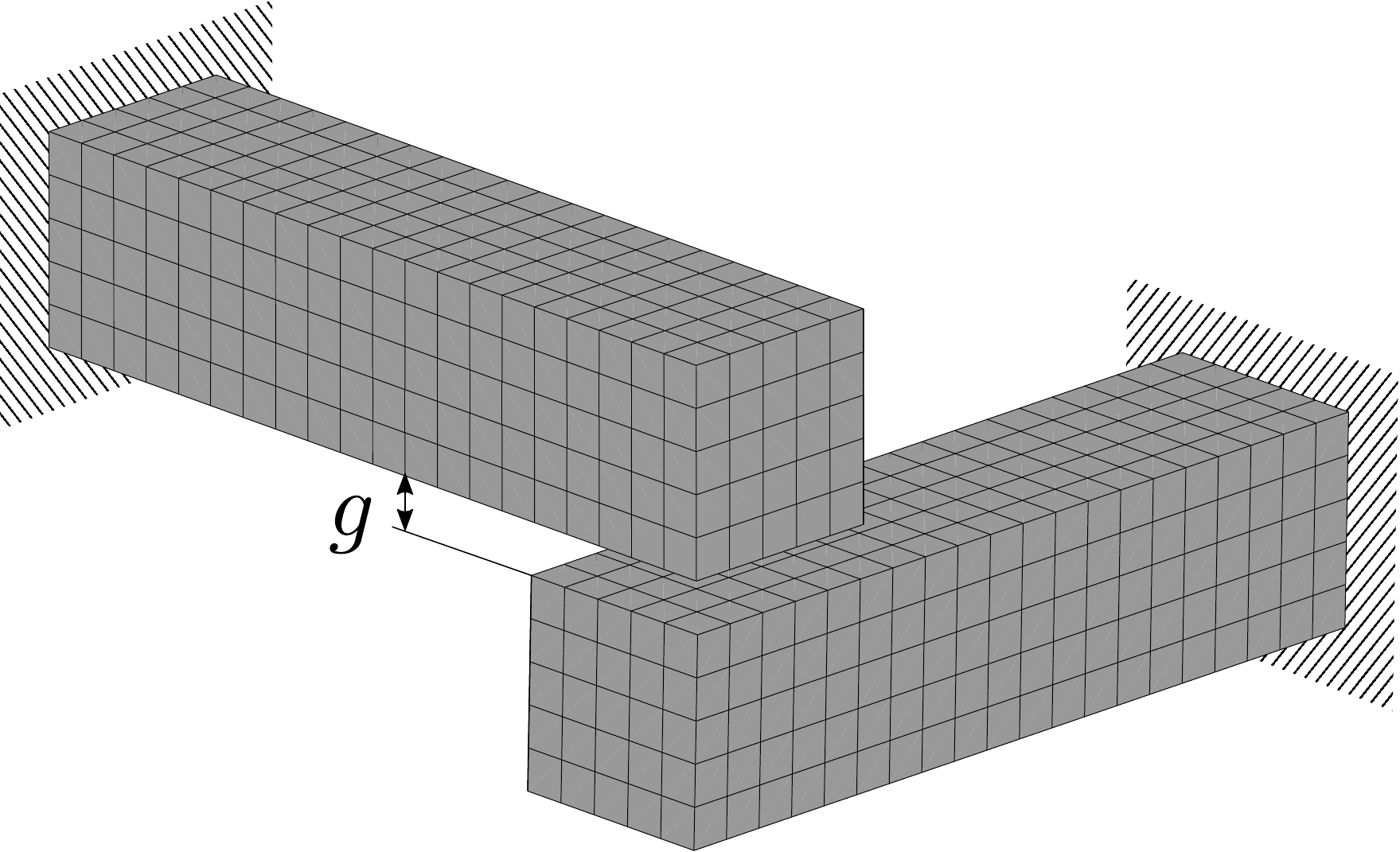}}
\caption{One single beam problem is analyzed from simulation data. The resulting system is composed of two of these beams interacting together.\label{fig:beam}}
\end{figure}

In this example we consider two viscoeleastic beams that can interact through contact between them, see Fig.\ref{fig:beam}, and whose physics are to be learned. Synthetic data come from finite element simulations, assuming a strain energy potential of the type
$$
U=C_{10}(\overline{I}_1-3)+C_{01}(\overline{I}_2-3)+\frac{1}{D_1}(J_{el}-1)^2,
$$
with $J_{el}$ the elastic volume ratio, $\overline{I}_1$ and $\overline{I}_2$ are the two invariants of the left Cauchy-Green deformation tensor, $C_{10}$ and $C_{01}$ are shear material constants and $D_1$ is the material compressibility parameter. The viscoelastic behavior is described by a two-term Prony series of the dimensionless shear relaxation modulus,
$$
g_R(t)=1-\bar{g}_1(1-e^{\frac{-t}{\tau_1}})-\bar{g}_2(1-e^{\frac{-t}{\tau_2}}),
$$
with relaxation coefficients of $\bar{g}_1$ and $\bar{g}_2$, and relaxation times of $\tau_1$ and $\tau_2$.

We assume that the necessary state variables for for a proper description of the beams are the position $\bs{q}$, its velocity $\bs{v}$ and the stress tensor $\bs{\sigma}$,
\begin{equation}\label{eq:beam_z}
\mathcal{S}=\{\bs{z}=(\bs{q},\bs{v},\bs{\sigma})\in\mathbb{R}^3\times\mathbb{R}^3\times\mathbb{R}^6\},
\end{equation}
at each node of the discretization of the beams. Since both beams are identical, see Fig. \ref{fig:beam} we characterize only one of them and develop a port-metriplectic learned simulator for the joint system. To do so, we employ thermodynamics-informed graph neural networks \cite{hernandez2022thermodynamics}.

Basically, a graph neural network is constructed on top of a graph structure $\mathcal{G}=(\mathcal{V},\mathcal{E},\bs{u})$, where $\mathcal{V}=\{1,...,n\}$ is a set of $\lvert\mathcal{V}\rvert=n$ vertices, $\mathcal{E}\subseteq\mathcal{V}\times\mathcal{V}$ is a set of $\lvert\mathcal{E}\rvert=e$ edges and $\bs{u}$ is the global feature vector. Each vertex and edge in the graph is associated with a node in the finite element model from which data are obtained. The global feature vector defines the properties shared by all the nodes in the graph, such as constitutive properties. More details on the precise formulation can be found at \cite{hernandez2022thermodynamics}.

To ensure traslational invariance of the learned model, the position variables of the system, $\bs{q}_i$, are assigned to the edge feature vector $\bs{e}_{ij}$ so the edge features represent relative distances ($\bs{q}_{ij}=\bs{q}_i-\bs{q}_j$) between nodes. The rest of the state variables are assigned to the node feature vector $\bs{v}_{i}$. We employ an encode-process-decode scheme \cite{battaglia2018relational}, built upon multilayer perceptrons (MLPs) shared between all the nodes and edges of the graph.

We use this graph-based framework to learn the self contribution of the dynamics, i.e. the first two terms of Eq. (\ref{eq:port}). The boundary terms are learned using a standard structure-preserving neural network \cite{hernandez2021structure} with the additional input of the external forces applied to the beam.

The dimensions of the beams are $H=10$, $W=10$ and $L=40$. The finite element mesh from which data are obtained consisted of $N_e=500$ hexahedral linear brick elements and $N=756$ nodes. The constitutive parameters are $C_{10}=1.5\nexp{5}$, $C_{01}=5\nexp{3}$, $D_1=10^{-7}$ and $\bar{g}_1=0.3$, $\bar{g}_2=0.49$, $\tau_1=0.2$, $\tau_2=0.5$ respectively. A distributed load of $F=10^5$ is applied in 52 different positions with an orientation perpendicular to the solid surface. Simulations were quasi-static and included $N_T=20$ time increments of $\Delta t=5\cdot 10^{-2}$ s. Two identical beams are assembled in $90^{\circ}$ with a gap of $g=10$, as depicted in Fig. \ref{fig:beam}.

\begin{figure}[h]
\centering
\pgfplotstableread{results/beam_box/error_beam_train.txt}\errortrain
\pgfplotstableread{results/beam_box/error_beam_test.txt}\errortest

\begin{tikzpicture}
\pgfplotsset{width=\textwidth, height=6cm}

  % q
  \begin{semilogyaxis}
  [xshift=0cm,
  area legend, boxplot/draw direction=y,
  grid=major, % Display a grid
  grid style={dashed,gray!30}, % Set the style
  cycle list={{red},{blue}},
  boxplot={draw position={1/3 + floor(\plotnumofactualtype/2) + 1/3*mod(\plotnumofactualtype,2)},box extend=0.3,},
  x=1cm,xtick={0,1,2,...,10},x tick label as interval,
  xticklabels={{$\bs{q}$},{$\bs{v}$},{$\bs{\sigma}$}},
  ticklabel style={font=\scriptsize},
  ylabel={Relative L2 error},
  custom legend,legend pos=outer north east,legend cell align=left,
  legend entries = {Train, Test},
  ] 
  	% q
  	% Train
	\addplot+[boxplot prepared from table={table=\errortrain,row=0,
    lower whisker=lw,
    upper whisker=uw,
    lower quartile=lq,
    upper quartile=uq,
    median=med}, boxplot prepared]
    coordinates {};
    % Test
	\addplot+[boxplot prepared from table={table=\errortest,row=0,
    lower whisker=lw,
    upper whisker=uw,
    lower quartile=lq,
    upper quartile=uq,
    median=med}, boxplot prepared] 
    coordinates {}; 
    
    % v
    % Train
	\addplot+[boxplot prepared from table={table=\errortrain,row=1,
    lower whisker=lw,
    upper whisker=uw,
    lower quartile=lq,
    upper quartile=uq,
    median=med}, boxplot prepared] 
    coordinates {}; 
    % Test
	\addplot+[boxplot prepared from table={table=\errortest,row=1,
    lower whisker=lw,
    upper whisker=uw,
    lower quartile=lq,
    upper quartile=uq,
    median=med}, boxplot prepared] 
    coordinates {}; 

    % sigma
    % Train
	\addplot+[boxplot prepared from table={table=\errortrain,row=2,
    lower whisker=lw,
    upper whisker=uw,
    lower quartile=lq,
    upper quartile=uq,
    median=med}, boxplot prepared] 
    coordinates {}; 
    \addlegendentry{~Train}
    % Test
	\addplot+[boxplot prepared from table={table=\errortest,row=2,
    lower whisker=lw,
    upper whisker=uw,
    lower quartile=lq,
    upper quartile=uq,
    median=med}, boxplot prepared] 
    coordinates {}; 
    \addlegendentry{~Test}

  \end{semilogyaxis}
  
\end{tikzpicture}
\caption{Box plots for the relative L2 error for all the rollout snapshots of the interacting beams in both train and test cases. The state variables represented are position ($\bs{q}$), velocity ($\bs{v}$), and stress tensor ($\bs{\sigma}$).}
\label{fig:beam_box}
\end{figure}
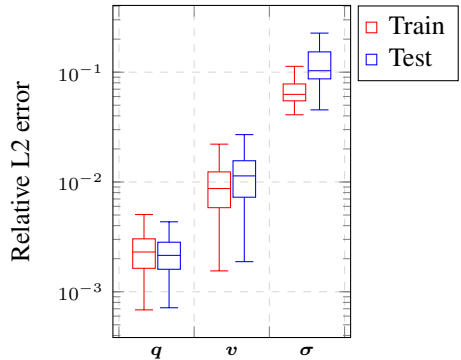

The results are presented in Fig. \ref{fig:beam_box}. The error magnitude is similar as the reported in previous work \cite{hernandez2022thermodynamics} in addition to the consistent formulation of post-metriplectic dynamics.

\section{Conclusions}\label{conc}

In this paper we have made two main contributions. On one side, the development of port-Hamiltonian-like approximations for dissipative open systems that communicate with other systems by exchanging energy and entropy through ports in their boundaries. This formulation extends the classical port-Hamiltonian approaches while guaranteeing the fulfillment of the laws of thermodynamics (conservation of energy in the bulk system, non-negative entropy production). The resulting formulation, which we refer as port-metriplectic---since it consists of a metric term and a symplectic one---, presents a rigorous thermodynamic description of the dissipative behavior of the system.

On the other hand, the just developed formulation is employed as an inductive bias for the machine learning of the physics of complex systems from measured data. This bias is developed as a soft constraint in the loss term, although it can also be imposed straightforwardly as a hard constraint.

The resulting neural networks, for which we have formulated two distinct versions, one based on standard multilayer perceptrons, and a second one based on graph neural networks, have shown an excellent performance. Error bars are equivalent to those obtained in previous works of the authors, by employing a closed-system approach to the same physics. The new approach opens the door to the development of learned simulators for complex systems through piece-wise learning of the physical behavior of each of its components. The final, global simulator is then obtained by assembling each piece through their ports.

\section*{Acknowledgements}

This material is based upon work supported in part by the Army Research Laboratory and the Army Research Office under contract/grant number W911NF2210271.

This work has also been partially funded by the Spanish Ministry of Science and Innovation, AEI /10.13039/501100011033, through Grant number PID2020-113463RB-C31. And by the \emph{Primeros Proyectos} Grant from Polytechnic University of Madrid, ETSII-UPM22-PM01.

The support of ESI Group through the Chairs at ENSAM Paris and Universidad de Zaragoza is also gratefully acknowledged.

\bibliography{PORT-METRIPLECTIC-ARXIV}
\bibliographystyle{unsrt}

%\end{multicols}

\end{document}